\documentclass[10pt,twocolumn,letterpaper]{article}

\usepackage[pagenumbers]{cvpr} 

\usepackage{graphicx}
\usepackage{amsmath}
\usepackage{amssymb}
\usepackage{booktabs}

\usepackage{setspace}
\usepackage{wrapfig}
\usepackage{comment}
\usepackage{color}
\usepackage{graphicx}
\usepackage{multirow}
\usepackage{tabularx}
\usepackage{bbding}
\usepackage{bbm}
\usepackage{float}
\usepackage{cite}
\usepackage{xspace}
\usepackage{colortbl}
\usepackage{makecell}
\usepackage[misc]{ifsym}
\usepackage{bbding}
\usepackage{pifont}
\usepackage{color,xcolor}
\usepackage{stfloats}
\usepackage{wasysym}
\newcommand{\cmark}{\ding{51}\xspace}%
\newcommand{\xmarkg}{\textcolor{lightgray}{\ding{55}}\xspace}%
\newcommand{\ours}{\textbf{D$^2$Zero}\xspace}

\usepackage[pagebackref,breaklinks,colorlinks]{hyperref}

\usepackage[capitalize]{cleveref}
\crefname{section}{Sec.}{Secs.}
\Crefname{section}{Section}{Sections}
\Crefname{table}{Table}{Tables}
\crefname{table}{Tab.}{Tabs.}

\def\cvprPaperID{956} 
\def\confName{CVPR}
\def\confYear{2023}

\begin{document}

\title{Semantic-Promoted Debiasing and Background Disambiguation\\for Zero-Shot Instance Segmentation}

\author{
Shuting He$^1$\footnotemark[2]
\qquad
Henghui Ding$^2$\footnotemark[2]~~$^{\textrm{\Letter}}$
\qquad
Wei Jiang$^1$\\
$^1$Zhejiang University
\qquad
$^2$Nanyang Technological University\\
\href{https://henghuiding.github.io/D2Zero}{https://henghuiding.github.io/D2Zero}
}
\maketitle
\renewcommand{\thefootnote}{\fnsymbol{footnote}}
\footnotetext[2]{Equal contribution.}
\footnotetext[0]{${\textrm{\Letter}}$ Corresponding author 
(henghui.ding@gmail.com).}

\begin{abstract}
   Zero-shot instance segmentation aims to detect and precisely segment objects of unseen categories without any training samples. Since the model is trained on seen categories, there is a strong bias that the model tends to classify all the objects into seen categories. Besides, there is a natural confusion between background and novel objects that have never shown up in training. These two challenges make novel objects hard to be raised in the final instance segmentation results. It is desired to rescue novel objects from background and dominated seen categories. To this end, we propose \ours~with Semantic-Promoted \underline{\textbf{D}}ebiasing and Background \underline{\textbf{D}}isambiguation to enhance the performance of \underline{\textbf{Zero}}-shot instance segmentation. Semantic-promoted debiasing utilizes inter-class semantic relationships to involve unseen categories in visual feature training and learns an input-conditional classifier to conduct dynamical classification based on the input image. Background disambiguation produces image-adaptive background representation to avoid mistaking novel objects for background. Extensive experiments show that we significantly outperform previous state-of-the-art methods by a large margin, \eg, \textbf{16.86\%} improvement on COCO.
   \vspace{-3mm}
\end{abstract}

\section{Introduction}
Existing fully supervised instance segmentation methods~\cite{maskrcnn,bolya2019yolact,solov1,TransformerSurvey} are commonly benchmarked on predefined datasets with an offline setting, where all categories are defined beforehand and learned at once, thus can neither handle novel concepts outside training datasets nor scale the model’s ability after training.
Perception errors inevitably arise when applying a trained instance segmentation model to scenarios that contain novel categories. 
To address these challenges, zero-shot instance segmentation (ZSIS)~\cite{zsi} is introduced to segment instances of unseen categories with no training images but semantic information only.

\begin{figure}[t]
    \centering
	\includegraphics[width=0.46\textwidth]{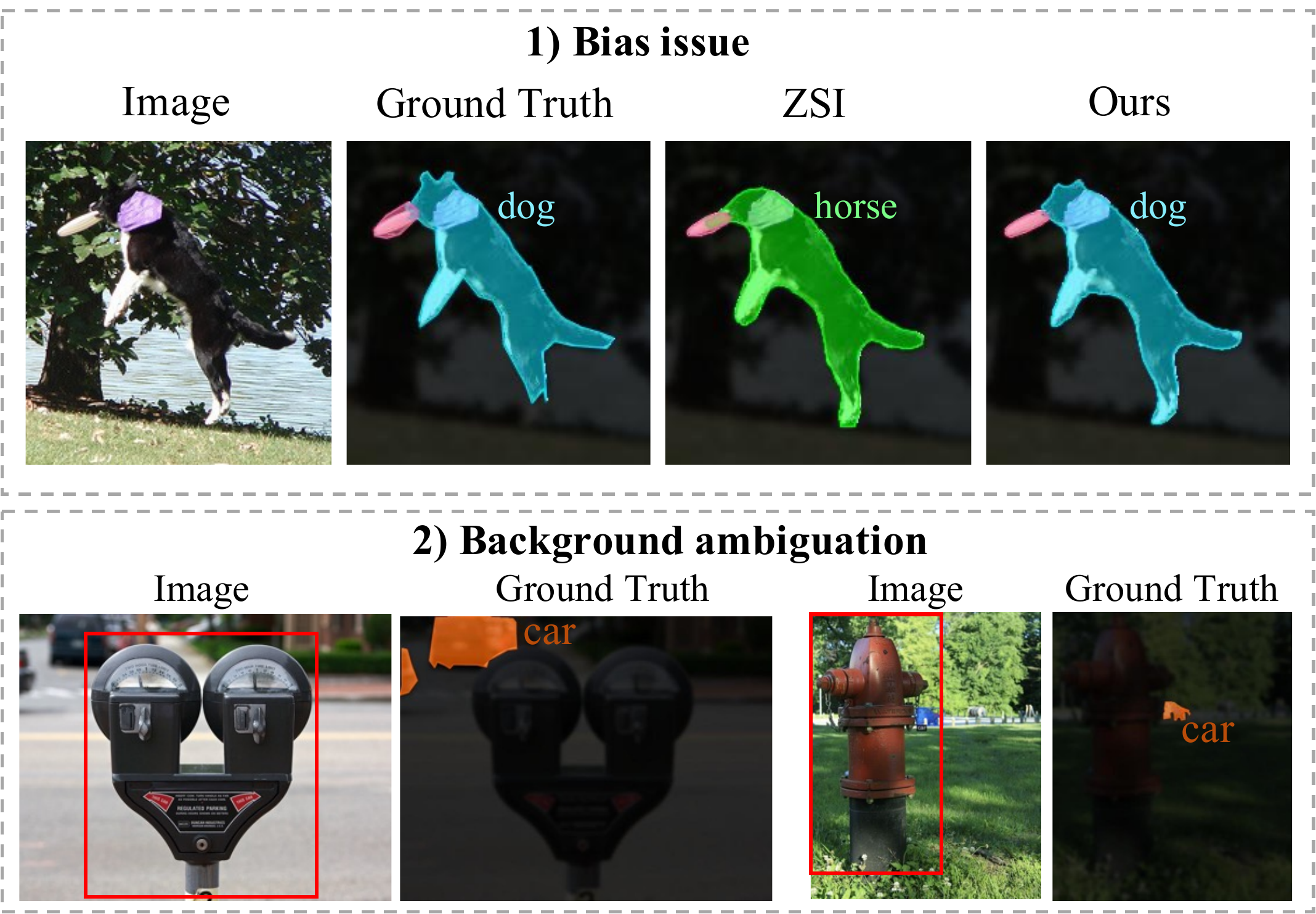}
	\vspace{-3mm}
	\caption{Two key challenges in generalized zero-shot instance segmentation.~1) Bias issue:~the model tends to label novel objects with seen categories, \eg, ZSI~\cite{zsi} incorrectly classifies unseen class \texttt{dog} as training class \texttt{horse}. 2) Background ambiguation: objects that do not belong to any training categories are considered background, \eg, \texttt{parking meter} and \texttt{fire hydrant}.} 
    \label{fig:teaser}
 \vspace{-2mm}
\end{figure}

Since scene images typically contain several objects of different categories, it is more realistic for ZSIS to segment both seen and unseen objects, which is termed Generalized ZSIS (GZSIS). In this work, we focus on two key challenges under GZSIS setting, bias issue and background ambiguation (see \figurename~\ref{fig:teaser}), and propose \textbf{D$^2$Zero} with semantic-promoted \underline{\textbf{D}}ebiasing and background \underline{\textbf{D}}isambiguation to enhance the performance of \underline{\textbf{Zero}}-shot instance segmentation.

Bias towards seen categories imposes a significant challenge to GZSIS. 
Since the model is trained on data of seen categories, it tends to classify all objects into seen categories, \eg, novel object \texttt{dog} is labeled as seen class \texttt{horse} in \figurename~\ref{fig:teaser}.
Previous work ZSI~\cite{zsi} introduces semantic embedding to build a mapping from seen classes to unseen ones then segments novel objects by sharing instance proposals of seen group and re-labeling these proposals within unseen group. Such a ``sharing'' strategy brings many false positives by assigning each instance two labels. Some zero-shot semantic segmentation methods~\cite{ZS3, CSRL, CaGNet} employ a generator to synthesize fake unseen features and fine-tune the classifier with these synthetic features. The generative way comes at the cost of forgetting some knowledge learned from seen categories and impairs the classifier's discriminative ability of the real feature. Besides, classifier is collapsed when a new class comes in, making the generative way impractical for application.
In this work, we address the bias issue from two aspects, feature extractor and classifier. 
Biased feature extractor mainly discriminate seen classes due to seen-only training objectives, 
which generalizes poorly to novel classes. We propose an unseen-constrained training objective to leverage semantic knowledge of unseen classes in visual feature learning. Specifically, we obtain semantic similarity of every seen-unseen class pair and generate a corresponding similarity-based pseudo unseen label for a seen object. 
Image features of seen classes are required to match the inter-class correlation with unseen classes under the supervision of pseudo unseen label, 
which enables the feature extractor to distinguish both seen and unseen classes.

Besides feature extractor, the bias devil also exists in the classifier. 
Previous zero-shot segmentation methods either use conventional fully-connected layer as classifier~\cite{ZS3,CSRL} or prototypical classifier built upon semantic embeddings~\cite{ding_iccv21,zsi}. 
However, these two types of classifier both have features clustered to fixed seen-class centers and do not consider the bias during inference. To address this issue, we design an input-conditional classifier based on transformer mechanism. We employ the semantic embeddings as query and visual features as key and value of transformer decoder, which bridges the semantic and visual spaces and transfers knowledge. 
Then the decoder outputs are employed as classifier in a prototypical way. The input-conditional classifier captures image-specific clues~\cite{lu2021simpler} and can better distinguish different categories of the input image. In such a way, the model learns to dynamically project semantic embeddings to input-conditional class centers, which greatly alleviates bias issue. Moreover, the input-conditional classifier establish the information interaction between visual and semantic spaces, contributing to mitigating multi-modal domain gap problem.

The background ambiguation issue is specific for zero-shot instance segmentation. In the training of instance segmentation, objects that do not belong to any training categories are considered background, \eg, \texttt{parking meter} and \texttt{hydrant} in \figurename~\ref{fig:teaser}. The model hence is likely to identify the novel objects as background, which affects the final performance a lot. 
To address this issue, BLC \cite{BLC} and ZSI \cite{zsi} propose to learn a background vector in the Region Proposal Network (RPN), which is optimized in a binary classifier of RPN. However, 
the binary classifier of RPN tends to overfit to seen categories and may fail to identify unseen categories~\cite{OLN, ovdetr}. We experimentally find that the Transformer~\cite{vaswani2017attention} based DETR-like model~\cite{detr,mask2former} can well generalize to novel categories in terms of proposal generation, thanks to its end-to-end training manner and classification-free instance proposal generation. 
Therefore, we collect all the foreground mask proposal produced by DETR-like model to get the global foreground mask and then apply the reverse of it on the feature map to get background prototype, which is used for background classification. 
Such an adaptive background prototype that updates according to input image can better capture image-specific and discriminative background visual clues, which helps to background disambiguation. 

Our main contributions are summarised as follows:
\vspace{-1mm}
\begin{itemize}
\setlength\itemsep{0em}
    \item We propose an unseen constrained visual feature learning strategy to leverage semantic knowledge of unseen categories in visual feature training, which facilitates mitigating bias issue in GZSIS.
    \vspace{-1mm}
    \item We design an input-conditional classifier that projects semantic embedding to image-specific visual prototypes, contributing to addressing both bias issue and multi-modal domain gap issue.
    \vspace{-1mm}
    \item To rescue novel objects from background, we introduce an image-adaptive background representation to better capture image-specific background clues.
    \vspace{-1mm}
    \item We achieve new state-of-the-art performance on zero-shot instance segmentation and significantly outperform ZSI~\cite{zsi} by a large margin, \eg, \textbf{16.86\%} HM-mAP under 48/17 split on COCO. 
    \vspace{-1mm}
\end{itemize}

\section{Related Work}
\textbf{Zero-Shot Image Classification} aims to classify images of unseen classes that have never shown up in training samples~\cite{lampert2009learning,palatucci2009zero,kankuekul2012online,lampert2013attribute,frome2013devise,hu2023suppressing}. 
There are two different settings: zero-shot learning (ZSL) and generalized zero-shot learning (GZSL). Under the ZSL setting~\cite{lampert2009learning,palatucci2009zero}, testing images are from unseen categories only. Typical ZSL methods include classifier-based way~\cite{akata2015evaluation,demirel2017attributes2classname,li2018deep} and instance-based way~\cite{socher2013zero,dinu2014improving,yu2013designing}, where the former one aims to learn a visual-semantic projection to transfer knowledge and the later one aims to synthesize fake unseen samples for training.
GZSL~\cite{scheirer2012toward} aims to identify samples of both seen and unseen categories simultaneously and suffers the challenge of a strong bias towards seen categories
\cite{chao2016empirical}. To address the bias issue, calibration methods~\cite{changpinyo2020classifier,das2019zero,guo2019dual} and detector-based methods~\cite{bhattacharjee2019autoencoder, felix2019generalised,socher2013zero} are introduced. The former way aims at calibrating the classification scores of seen categories to achieve a trade-off balance between seen and unseen groups, while the detector-based way explores identifying the unseen samples as out-of-distribution and classifying these unseen samples within unseen categories.

\textbf{Zero-Shot Instance Segmentation (ZSIS)}. Fully supervised instance segmentation are extensively studied in recent years~\cite{maskrcnn,bolya2019yolact,solov1}, which however are data-driven and cannot handle unseen classes that have never shown up in training. Recently, zero-shot instance segmentation is raised by ZSI~\cite{zsi} to apply zero-shot learning to instance segmentation. There are two test settings: zero-shot instance segmentation (ZSIS) and generalized zero-shot instance segmentation (GZSIS), where GZSIS is more realistic since an image typically contains multiple objects of different seen/unseen categories. In this work, we mainly focus on GZSIS and address its two key challenges, bias issue, and background confusion. ZSI~\cite{zsi} addresses the bias issue by copying all the instances detected as seen categories and re-label these instances within unseen group, resulting in many false positives. In this work, we propose an unseen-constrained visual training strategy and input-conditional classifier to alleviate the bias issue.

\textbf{Zero-Shot Semantic Segmentation (ZSSS)}~\cite{ZS3, SPNet, FZShot3D} aims to segment the image to semantic regions~\cite{ding2018context} of seen and unseen categories, it shares some commonalities with ZSIS. Existing ZSSS methods can be divided into two ways: embedding-based methods and generative-based methods. Embedding-based methods~\cite{pastore2021closer,SPNet,ZegFormer,zsseg_baseline, lv2020learning, ding_iccv21, hu2020uncertainty} project visual and semantic features to a common space, \eg, semantic, visual, or latent space, to transfer knowledge and conduct classification in this common space. 
Generative-based ZSSS methods~\cite{CSRL,ZS3,SIGN,PADing} utilize a feature generator to synthesize fake features for unseen categories.

\textbf{Language-driven Segmentation} shares some similarities with ZSIS. They utilize language information to guide the segmentation, \eg, referring expression segmentation \cite{VLT,ding2022vlt,liu2022instance,yang2022lavt,M3Att,GRES} and open-vocabulary segmentation \cite{huynh2022open,ghiasi2022scaling,li2022languagedriven,wu2023betrayed}. However, instead of following the strict zero-shot setting of excluding any unseen classes in training data, these works allow as many classes as possible to implicitly participate in model training by using image captions or referring expressions, which is however considered as information leakage in the zero-shot learning setting.

\section{Approach}

\begin{figure*}[t]
    \centering
	\includegraphics[width=1.0\textwidth]{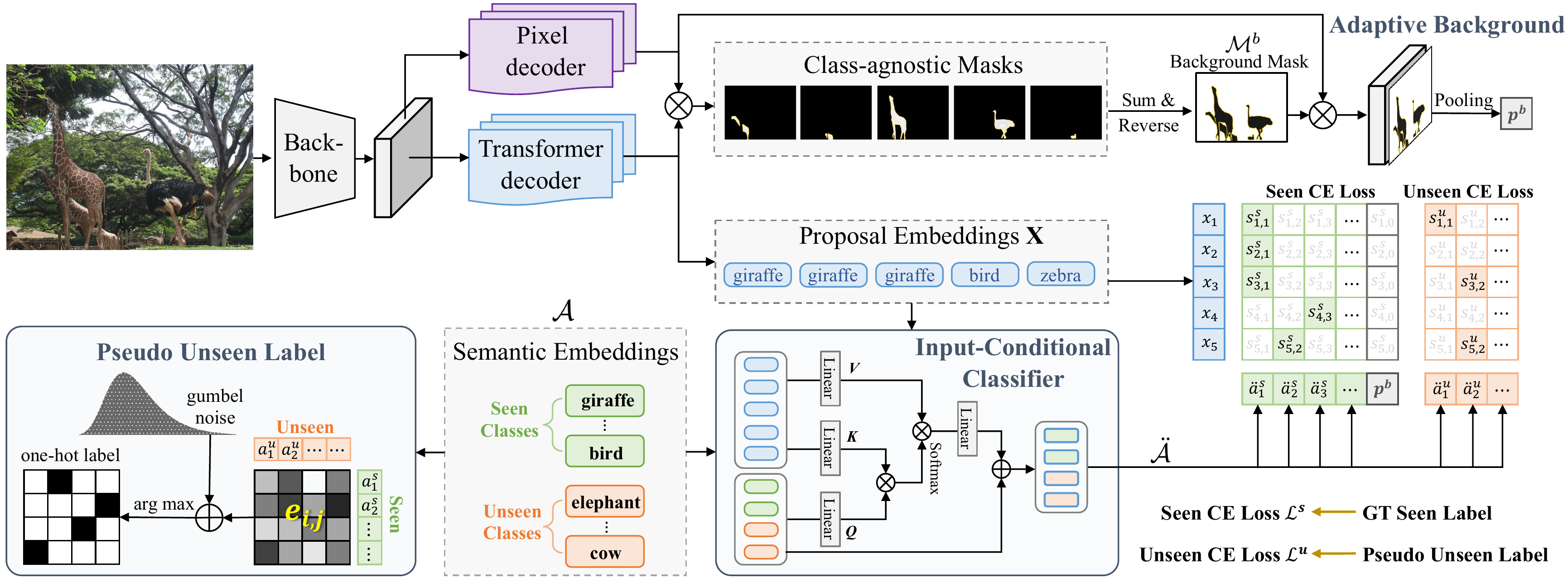}
	\vspace{-0.66cm}
	\caption{Framework overview of our \ours. The model proposes a set of class-agnostic masks and their corresponding proposal embeddings. The proposed input-conditional classifier takes semantic embeddings and proposal embeddings as input and generates image-specific prototypes. Then we use these prototypes to classify image embeddings, under the supervision of both seen CE loss $\mathcal{L}^s$ and unseen CE loss $\mathcal{L}^u$. The unseen CE loss enables unseen classes to join the training of feature extractor. We collect all the masks and produce a background mask, then apply this mask to the image feature to generate an image-adaptive background prototype for classification.} 
	\vspace{-2mm}
	\label{fig:framework}
\end{figure*}
\subsection{Problem Formulation}
In zero-shot instance segmentation, there are two non-overlapping foreground groups, $N^s$ seen categories denoted as $C^s$ and $N^u$ unseen categories denoted as $C^u$, and a background class $c^b$, where $C^s$ $=$ $\{c^s_1,c^s_2,...,c^s_{N^s}\}$ and  $C^u$ $=$ $\{c^u_1,c^u_2,...,c^u_{N^u}\}$. Each category has a corresponding semantic embedding, denoted as $\mathcal{A}$ $=$ $\{\mathbf{a}^b,\mathbf{a}^s_1,\mathbf{a}^s_2,..,\mathbf{a}^s_{N^s},\mathbf{a}^u_1,\mathbf{a}^u_2,...,\mathbf{a}^u_{N^u}\}$. Given an image set that contains $N^i$ images of $N^s$ and $N^u$ categories, the training set $D_{train}$ is built from training images of seen categories, \ie, each training image that contains any objects of $\{c^s_1,c^s_2,...,c^s_{N^s}\}$ but no object of unseen categories. According to whether considering seen classes during inference, there are two different settings, one is ZSIS which segments objects of only the unseen categories and the other is Generalized ZSIS (GZSIS) which segments objects of both seen and unseen categories. GZSIS is more realistic since an image usually includes multiple objects and we cannot ensure there is no object of seen categories.

\subsection{Architecture Overview}
The architecture overview of our proposed D$^2$Zero is shown in \figurename~\ref{fig:framework}. We adopt ResNet-50~\cite{he2016deep} as backbone and follow the paradigm of Mask2Former~\cite{mask2former}. 
Mask2Former seamlessly converts pixel classification to mask classification with careful design of proposal embeddings $\{\mathbf{x}_n\}_{n=1}^{N^p}$$\in$$\mathbb{R}^{d}$ and mask predictions $\{\mathcal{M}_n\}_{n=1}^{N^p}$$\in$$\mathbb{R}^{H\times W}$. Since the masks are class-agnostic, the model is endowed with the ability to generate masks for novel objects that have never shown up in training set~\cite{ovdetr}. We generate a set of prototypes as input-conditional classifier. The background prototype is generated by masked average pooling on the image feature, where the background region is decided by all the class-agnostic masks. Then we adopt seen cross-entropy loss and our proposed unseen cross-entropy loss as training objectives.

\subsection{Semantic-Promoted Visual Feature Debiasing}
Due to the lack of unseen categories' training data, the model is trained on samples of seen categories only. As a consequence, there is a strong bias towards seen categories that the model tends to classify all the testing objects into seen categories~\cite{chao2016empirical}. To address the bias issue, ZSI~\cite{zsi} separates the classification of seen and unseen categories and labels each instance with two labels, one from seen group and the other from unseen group. This strategy, though works, sidesteps the essence of the bias problem. In this work, we explore alleviating the bias issue in zero-shot instance segmentation and attribute it to 1) biased feature extractor that focuses on producing features to discriminate seen categories and 2) biased classifier that tends to capture clues derived from training data statistics. We herein propose an unseen-constrained feature extractor and input-conditional classifier to address the biased feature extractor and biased classifier, respectively. The unseen-constrained feature extractor utilizes inter-class semantic similarities to guide the training of visual feature extractor, in which seen-unseen relationships are involved as training objective thus unseen categories can join in the visual feature training. The input-conditional classifier learns a semantic-visual alignment based on transformer and generates input-specific visual representations as classifier prototypes.

\vspace{2mm}
\noindent\textbf{Unseen-Constrained Visual Feature Learning}
\vspace{1mm}

The feature extractor trained on seen categories focuses more on features useful to discriminate seen classes, inducing a loss of information required to deal with unseen classes. To address this issue and produce features that generalize better to novel concepts, we propose to introduce semantic information of unseen categories as training guidance to constrain visual feature learning. We first generate an inter-class correlation coefficient by calculating the semantic similarity of every unseen-seen category pair,
\vspace{-0.75mm}\begin{equation}\vspace{-0.75mm}
    e_{i,j} = \frac{\mathrm{exp}(<\mathbf{a}^s_i,\mathbf{a}^u_j>/\tau)}{\sum_{k=1}^{N^u}\mathrm{exp}(<\mathbf{a}^s_i,\mathbf{a}^u_k>/\tau)},
\end{equation}
where $<,>$ is cosine similarity, $\tau$ is the temperature parameter, $e_{i,j}$ is a soft value in the range of $[0,1]$ and represents correlation coefficient of the $i$-th seen embedding $\mathbf{a}_i^s$ and the $j$-th unseen embedding $\mathbf{a}_j^u$, the higher $e_{i,j}$ represents the closer relationship. For each of the $N^s$ seen categories, there are $N^u$ coefficients, \ie, $\mathbf{e}_i=\{e_{i,j}\}_{j=1}^{N^u}$.
The inter-class correlation matrix prior is then used to guide the visual feature learning. Instead of using original soft probability, we choose a pseudo unseen label for each seen object based on the coefficient $e_{i,j}$. Specifically, we employ Gumbel-Softmax trick~\cite{GumbelSoftmax} to form a Gumbel-Softmax distribution and transform $\mathbf{e}_i$ to discrete variable $\dot{\mathbf{e}}_i\in\{0,1\}^{N^u}$
\vspace{-0.75mm}\begin{equation}\vspace{-0.75mm}
    \dot{\mathbf{e}}_i = \mathrm{onehot}\Big(\mathop{\arg\max}_{j} [g_j+{e_{i,j}}]\Big),
\end{equation}
where $g_1,...,g_{N^u}$ are random noise samples drawn from Gumbel $(0,1)$ distribution. The pseudo unseen label $\dot{\mathbf{e}}_i$ changes with training iterations, following the rule that the larger $e_{i,j}$ has the higher probability of $c^u_j$ being chosen as the pseudo unseen label of $c^s_i$. For each proposal embedding $\mathbf{x}_n$, a classification score $\mathbf{s}_{n}^u$ of unseen group is obtained by
\vspace{-0.75mm}\begin{equation}\vspace{-0.75mm}
    {s}_{n,j}^u = \frac{\mathrm{exp}(\mathrm{MLP}(\mathbf{x}_n)\mathbf{a}_j^u/\tau)}{\sum_{k=1}^{N^u}\mathrm{exp}(\mathrm{MLP}(\mathbf{x}_n)\mathbf{a}_k^u/\tau)},
\label{Eq:unseen_score}
\end{equation}
where $\mathbf{s}_{n}^u=\{{s}_{n,j}^u\}_{j=1}^{N^u}\in\mathbb{R}^{{N^u}}$, $\mathrm{MLP}$ denotes Multi-layer Perceptron. The ground truth of $\mathbf{s}_{n}^u$ is $\dot{\mathbf{e}}_{c_n}$, where $c_n$ denotes the ground truth seen label index of $\mathbf{x}_n$. The unseen cross-entropy loss $\mathcal{L}^u$ is applied on $\mathbf{s}_n^u$ to have the model learn pseudo classification among unseen categories,
\vspace{-1.25mm}\begin{equation}\vspace{-1.25mm}
    \mathcal{L}^u = -\frac{1}{N^f}\sum_{n=1}^{N^f}\sum_{j=1}^{N^u}\dot{e}_{c_n,j}log{s}_{n,j}^u,
\label{Eq:unseen_ce}
\end{equation}
where $N^f$ is the number of proposals with foreground labels. It's worth noting that Eq.~(\ref{Eq:unseen_ce}) is applied on $\mathbf{x}_n$ of foreground objects while disabled for \texttt{background}.

Meantime, for each proposal embedding $\mathbf{x}_n$, a classification score $\mathbf{s}_n^s$ of seen group is obtained by
\vspace{-0.75mm}\begin{equation}\vspace{-0.75mm}
    {s}_{n,i}^s = \frac{\mathrm{exp}(\mathbf{x}_n\mathbf{a}_i^s/\tau)}{\sum_{k=0}^{N^s}\mathrm{exp}(\mathbf{x}_n\mathbf{a}_k^s/\tau)},
\label{Eq:seen_score}
\end{equation}
where $\mathbf{s}_n^s=\{s_{n,i}^s\}_{i=0}^{N^s}\in\mathbb{R}^{({N^s+1})}$ is the classification score for the $n$-th proposal, $\mathbf{a}_0^s=\mathbf{a}^b$ and $s_{n,0}^s$ represents score of \texttt{background}. A cross-entropy loss is applied on $\mathbf{s}_n^s$ to guide the classification among seen categories,
\vspace{-1mm}\begin{equation}\vspace{-1mm}
    \mathcal{L}^s = -\frac{1}{N^p}\sum_{n=1}^{N^p}\sum_{i=0}^{N^s}\mathbbm{1}(c_n=i)log{s}_{n,i}^s,
\end{equation}
where $\mathbbm{1}(*)$ outputs 1 when ${*}$ is true otherwise 0, $c_n$ is the ground truth label of $n$-th proposal. $N^p$ is the number of proposals. The overall training objective is $\mathcal{L} = \mathcal{L}^s+\lambda\mathcal{L}^u$.

With the unseen cross-entropy loss $\mathcal{L}^u$, the feature extractor is also trained under the constraints of unseen categories instead of only under the constraints of seen categories, which greatly help the feature extractor capture clues that are useful for unseen categories.

\vspace{3mm}
\noindent\textbf{Input-Conditional Classifier}
\vspace{1mm}

Directly using semantic embeddings $\mathcal{A}$ as classifier, though helps to semantically links knowledge of seen and unseen groups, makes the features clustered to fixed class centers and does not consider the bias issue in classifier. To further alleviate the bias issue in zero-shot instance segmentation, we propose an input-conditional classifier that dynamically classifies visual embeddings according to input features. As shown in \figurename~\ref{fig:framework}, semantic embeddings $\mathbf{a}_i$ are employed as query $Q$ in a transformer module, while key $K$ and value $V$ are concatenation of proposal embeddings, \ie, $\mathbf{X}=[\mathbf{x}_1, \mathbf{x}_2,...,\mathbf{x}_{N^p}]$, where $[,]$ denotes concatenation operation. After transformer module, semantic-projected visual embeddings $\ddot{\mathbf{a}}_i$ that are conditional on $\mathbf{x}_n$ are generated. In detail, given semantic embeddings $\mathcal{A}=[\mathbf{a}^s_1,\mathbf{a}^s_2,..,\mathbf{a}^s_{N^s},\mathbf{a}^u_1,\mathbf{a}^u_2,...,\mathbf{a}^u_{N^u}]$, a self-attention is first performed on $\mathcal{A}^s$ and outputs $\hat{\mathcal{A}}^s$. Then cross-attention is performed as
\vspace{-2mm}\begin{equation}\vspace{-2mm}
\begin{aligned}
    Q&=\mathbf{w}_Q\hat{\mathcal{A}},\quad K=\mathbf{w}_K\mathbf{X},\quad  V=\mathbf{w}_V\mathbf{X},\\
    \ddot{\mathcal{A}}&= \mathrm{MHA}(Q,\, K,\, V) = \mathrm{softmax}(\frac{QK^{T}}{\sqrt{d_k}})V,
\end{aligned}
\end{equation}
where $\mathbf{w}_Q,\mathbf{w}_K,\mathbf{w}_V$ are learnable parameters of three independent linear layers mapping inputs to the same intermediate representations of dimension $d_k$. $\ddot{\mathcal{A}}^s$ is the desired image-specific semantic-visual embedding. We then update the classifier by replacing the original semantic embedding $\mathbf{a}^s_j$/$\mathbf{a}^s_k$ in Eq.~(\ref{Eq:seen_score}) and $\mathbf{a}^u_i$/$\mathbf{a}^u_k$ in Eq.~(\ref{Eq:unseen_score}) to input-conditional semantic embedding $\ddot{\mathbf{a}}^s_i$/$\ddot{\mathbf{a}}^s_k$ and $\ddot{\mathbf{a}}^u_j$/$\ddot{\mathbf{a}}^u_k$, respectively. 

$\ddot{\mathcal{A}}$ has three main advantages over original semantic embedding ${\mathcal{A}}$. First, $\ddot{\mathcal{A}}$ is projected from semantic space to visual space via interaction with visual proposal embedding $\mathbf{X}$, which helps to mitigate visual-semantic domain gap and makes the classification easier to be learned. Second, $\ddot{\mathcal{A}}$ capture image-specific clues according to input feature and can better adaptively distinguish different categories of the input image. What's more, the class centers by $\ddot{\mathcal{A}}$ are input-conditional instead of fixed, thus the visual features trained with such dynamic classifier would not collapse to several fixed feature centers but tend to capture discriminative inter-class distance, which greatly helps to mitigate bias issue.

\subsection{Image-Adaptive Background Disambiguation }
There is confusion between background and unseen objects in zero-shot instance segmentation. The unseen categories do not join the training of segmentation model, which is trained to identify objects of seen categories as foreground objects and others as background, so they are easy to be mistaken for background. ZSI~\cite{zsi} argues that the semantic word ``\texttt{background}'' cannot represent background class and propose Background Aware RPN (BA-RPN) \& Synchronized Background to use a vector learned in RPN as background representation in zero-shot classifier. However, this learned vector is fixed after training and cannot be changed according to the input image, which limits its representation to complex backgrounds and generalization ability to novel scenarios. This background parameter is optimized in a binary classifier of RPN, which tends to overfit to seen categories and may fail to identify unseen categories~\cite{OLN, ovdetr}. To address this issue, we herein propose an image-adaptive background disambiguation that adaptively generates high-quality background representation according to the input image.

Specifically, we gather all the proposed binary masks $\{\mathcal{M}_n\}_{n=1}^{N^p}$ obtained from our model to indicate foreground region $\mathcal{M}^f$, \ie, $\mathcal{M}^f_{(x,y)}\!=\!\mathrm{max}(\mathcal{M}_{0,(x,y)},...,\mathcal{M}_{{N^p},(x,y)},)$, where $(x,y)$ denotes pixel position and $\mathcal{M}^f_{(x,y)}\!=\!1$ represents that the pixel $(x,y)$ belongs to foreground. It's worth noting that we gather all the proposed masks to ensure a high recall of foreground region, which is desired to detect novel objects. The background mask $\mathcal{M}^b$ is generated by taking the reverse of foreground mask, $\mathcal{M}^b=1-\mathcal{M}^f$. Then a Mask Average Pooling (MAP) is performed on visual feature maps to get background prototype,
\vspace{-1mm}\begin{equation}\vspace{-1mm}
    \mathbf{p}^b = \frac{\sum_{(x,y)} \mathcal{M}^b_{(x,y)}\mathbf{F}_{(x,y)}}{\sum_{(x,y)}\mathcal{M}^b_{(x,y)}}.
\end{equation}
We use this prototype to replace $\mathbf{a}_0^s$ Eq.~(\ref{Eq:seen_score}). $\mathbf{p}^b$ is adaptive according to visual feature and thus can better capture image-specific and discriminative background visual clues.

\textbf{Comparison with word embedding background and learned-parameter background}. \textbf{1)} Word embedding of background is either learned from large-scale text data without seeing visual data, \eg, word2vec~\cite{word2vector}, or derived from text encoder trained on large-scale text-image pairs of ``thing'' classes, \eg, CLIP~\cite{CLIP}. Thus, the existing background word-vector cannot well represent the complex visual appearance of background. \textbf{2)} ZSI~\cite{zsi} learns a background vector in the Region Proposal Network (RPN) and uses this vector to update the semantic embedding of background class. Such a learned vector is optimized in a binary classifier of RPN and captures some visual patterns. However, it is fixed after training and may identify novel objects as background, since the binary classifier of RPN tends to overfit to seen categories and may fail to identify unseen categories~\cite{OLN, ovdetr}. \textbf{3)} Our proposed image-adaptive prototype is visual feature obtained from the image background region directly and captures more useful visual clues. Compared to BA-RPN of ZSI~\cite{zsi} using a binary classifier, our DETR-like model can better generalize to novel categories in terms of proposing foreground instances because of its classification-free instance proposal manner. The proposed adaptive background prototype changes according to the input image and can better capture image-specific and discriminative background visual clues.

\section{Experiments}

\subsection{Experimental Setup}
\noindent\textbf{Implementation Details.} The proposed approach is implemented with the public platform Pytorch. We use ResNet-50 \cite{he2016deep} based Mask2Former~\cite{mask2former} to generate class-agnostic masks and corresponding proposal embeddings. All hyper-parameters are consistent with the default settings unless otherwise specified. We use CLIP~\cite{CLIP} to extract semantic embeddings of COCO classes. Meantime, for a fair comparison with previous works, we also report our results based on word2vec~\cite{mikolov2013distributed}. 
Hyper-parameter $\lambda$ and $\tau$ are set to 0.1, 0.1, respectively. 
The model is optimized using Adamw with learning rate set to 0.0001, trained on 8 RTX2080Ti~\!(12G) with batch size set to 16.

\vspace{1mm}
\noindent\textbf{Dataset \& Training/Testing Setting.} Following ZSI~\cite{zsi}, we use MS-COCO 2014~\cite{ms_coco} instance segmentation dataset containing 80 classes to train and evaluate our proposed approach. 
Two different splits of seen and unseen categories are built to evaluate zero-shot ability. The first is 48/17 split with 48 seen categories and 17 unseen categories. The second is 65/15 split with 65 seen categories and 15 unseen categories. Training set is built from images containing seen categories only. To avoid information leakage, the images that contain any pixels of unseen categories are removed from training set, which is different from open-vocabulary setting that allows using of some unseen images~\cite{ovrcnn}. In testing set, all the MS-COCO testing images that contain pixels of unseen categories are selected. 

\vspace{1mm}
\noindent\textbf{Metrics.} Following ZSI~\cite{zsi}, Recall@100, \ie, top 100 instances, with IoU thresholds of $\{0.4,0.5,0.6\}$ and mean Average Precision (mAP) with IoU thresholds of 0.5 are employed to report the performance. Under GZSIS setting, seen categories far outperform unseen categories and overmaster the Recall@100 and mAP. To better reveal unseen categories' effects on overall performance, we compute the harmonic mean (HM)~\cite{xian2018zero} of seen and unseen categories, where $\mathrm{HM(A,B)=2AB/(A+B)}$.

\vspace{1mm}
\noindent\textbf{Text Prompts.} We follow previous works~\cite{VILD, CLIP} to generate the text embeddings using prompt ensembling. For each category, we utilize multiple prompt templates and then obtain the final text embeddings via averaging.

\vspace{-2mm}
\subsection{Component Analysis}\vspace{-2mm}
We conduct extensive experiments to verify the effectiveness of our proposed components in \tablename~\ref{tab:ablation_study} with both 48/17 and 65/15 splits. 
We design our baseline by replacing the learnable classifier with text embeddings to classify image embeddings, which is similar to VILD-Text~\cite{VILD}. As we can see, there is a serious bias towards seen categories issue, \eg, unseen mAP 7.15\% is much lower than seen mAP 53.49\%.
In the following, we analyze our proposed component from a qualitative and quantitative perspective. We get the final results by combining all the components, which significantly surpass our baseline. 
\begin{table}[t]
\setlength\tabcolsep{5pt}
\scriptsize
  \begin{center}
  \begin{tabular}{ccclcccccc}
  \toprule
   &  & &  &\multicolumn{2}{c}{Seen} & \multicolumn{2}{c}{Unseen} & \multicolumn{2}{c}{HM}\\
Split  & $\ddot{\mathcal{A}}$ & $\mathbf{p}^b$ & $\mathcal{L}^u$ &  mAP  & Recall & mAP  & Recall & mAP & Recall \\
  \hline
\multirow{5}{*}{48/17} & \xmarkg & \xmarkg & \xmarkg& 53.49   &  \textbf{77.52}  &\ \  7.15   &     32.39 &   12.61  & 45.68 \\
& \cmark& \xmarkg& \xmarkg &   53.24 & 76.11   & 11.95    &   36.53  & 19.52    & 49.36 \\
& \xmarkg& \cmark& \xmarkg &  53.17  &  76.13  &  10.06   &   36.71  & 16.91    & 49.53 \\
& \xmarkg& \xmarkg& \cmark &  52.78  &  75.69  &  11.34   &  38.23   &  18.66   & 50.80 \\
& \cmark& \cmark& \cmark &   \textbf{54.42} &   {76.22} &  \textbf{15.06}   &  \textbf{38.38}   & \textbf{23.59}    &  \textbf{51.06}\\
  \arrayrulecolor{gray}\hline
\multirow{5}{*}{65/15} & \xmarkg & \xmarkg & \xmarkg&  40.64  & 74.91   &  15.65   &  35.61   &  22.59   & 48.27 \\
& \cmark& \xmarkg& \xmarkg &    \textbf{41.26}  & \textbf{75.41}   &  18.89   &  40.64   &   25.91  & 52.82  \\
& \xmarkg& \cmark& \xmarkg &   40.45 &  74.18  &  17.78   & 38.12    &  24.70   & 50.36 \\
& \xmarkg& \xmarkg& \cmark &  39.51 &  73.87  &  18.23   &   41.38  &   24.94  & 53.04\\
& \cmark& \cmark& \cmark &    {41.18}  &{74.94}  &\textbf{20.22}  &\textbf{46.01}  & \textbf{ 27.13} &\textbf{57.01}  \\
 \arrayrulecolor{black}\bottomrule
  \end{tabular}
\vspace{-3mm}
\caption{Component Analysis of our \ours~under GZSIS setting. $\ddot{\mathcal{A}}$ represents input-conditional classifier. $\mathbf{p}^b$ is image-adaptive background. $\mathcal{L}^u$ represents unseen CE loss.}\label{tab:ablation_study}
\vspace{-5mm}
\end{center}
\end{table}

\vspace{1mm}
\noindent\textbf{Input-Conditional Classifier.}
By replacing the conventional text embedding classifier with our input-conditional classifier, we can obtain significant improvement on unseen results, \eg, $\uparrow$4.8\% mAP on 48/17 split. The improvement on unseen group brings performance gain to HM results, \eg, 6.91\% HM-mAP and 3.68\% HM-Recall gains on 48/17. Owing to our delicate design of classifier, the issues of bias toward seen categories and domain gap are greatly alleviated. Such significant improvements validate the superiority of our input-conditional classifier quantitatively.

\begin{figure}[t]
    \centering
	\includegraphics[width=0.45\textwidth]{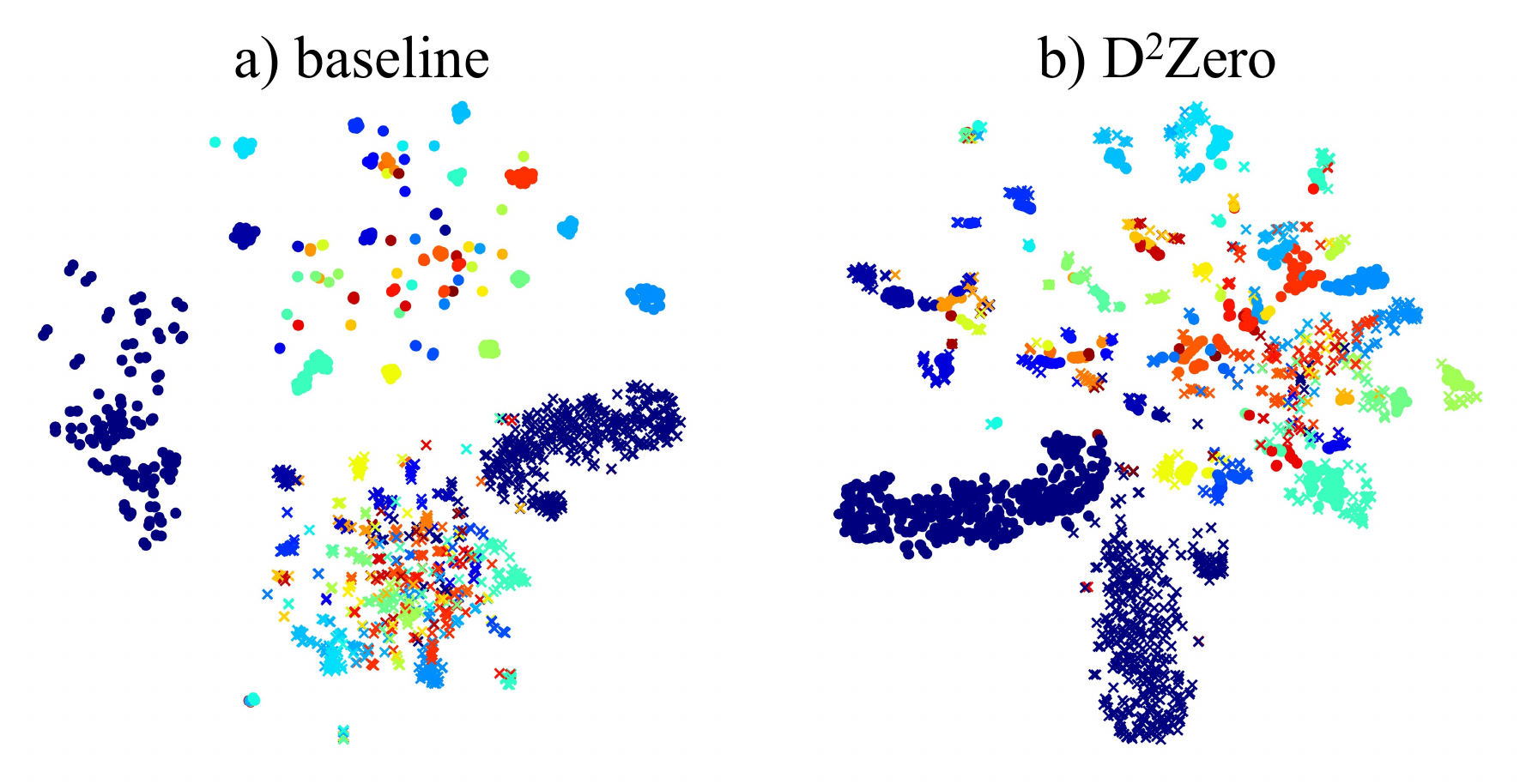}
	\vspace{-0.36cm}
	\caption{(Best viewed in color) t-SNE~\cite{tsne} visualization of image and text embeddings distribution on 48/17 split. The circle denotes the image embeddings. The cross denotes the text embeddings. Samples from different classes are marked in different colors.} 
	\vspace{-0.3cm}
	\label{fig:tsne}
\end{figure}

In \figurename~\ref{fig:tsne}, we utilize t-SNE~\cite{tsne} to visualize the image and text embeddings distribution with and without our input-conditional classifier. The t-SNE samples are from the same image with 4 classes. As shown in \figurename~\ref{fig:tsne}(a), image embeddings (circles) and corresponding text embeddings (cross) are far away from each other, because of the domain gap between vision and language.
In \figurename~\ref{fig:tsne}(b), cross-modal features from same class, \eg, the yellow circles and yellow cross, are pulled closer, showing
high intra-class compactness and inter-class separability characteristics.
Both quantitative and qualitative results demonstrate that with input-conditional classifier, image embeddings are well aligned with text embeddings and are capable of capturing discriminative features.

\begin{table}[t]
\setlength\tabcolsep{5pt}
\scriptsize
  \begin{center}
  \begin{tabular}{l|cccccc}
  \toprule
  \multirow{2}{*}{{Method}}&\multicolumn{2}{c}{Seen} & \multicolumn{2}{c}{Unseen} & \multicolumn{2}{c}{HM}\\
 &  mAP  & Recall & mAP  & Recall & mAP & Recall \\
  \hline
word embedding bg &  \textbf{53.49}   &  \textbf{77.52}  &\ \  7.15   &     32.39 &   12.61  & 45.68 \\
learned-parameter bg &  53.01   & 76.21   &\ \  8.74  &    34.41  &  15.01   & 47.41\\
\rowcolor{lightgray!30}\ours~bg &  {53.17} &76.13 & \textbf{10.06} & \textbf{36.71}&  \textbf{16.91}&  \textbf{49.53} \\
 \arrayrulecolor{black}\bottomrule
  \end{tabular}
\vspace{-3mm}
\caption{Comparison of different background (bg) designs.}\label{tab:background}
\vspace{-5mm}
\end{center}
\end{table}

\vspace{1mm}
\noindent\textbf{Image-Adaptive Background.}
The different choices of background design have great impacts on the final performance, as shown in \tablename~\ref{tab:background}. The experiments are performed on our baseline and under 48/17 split. Learned-parameter bg surpasses word embedding bg with 2.40\% HM-mAP and 1.73\% HM-Recall, respectively, which indicates that learnable bg can mitigate background ambiguation to some extent.
Compared with our image-adaptive background, using word embedding bg or learned-parameter bg, both mAP and Recall suffer from degradation, which verifies the effectiveness of our proposed approach.

\begin{figure}[t]
    \centering
	\includegraphics[width=0.46\textwidth]{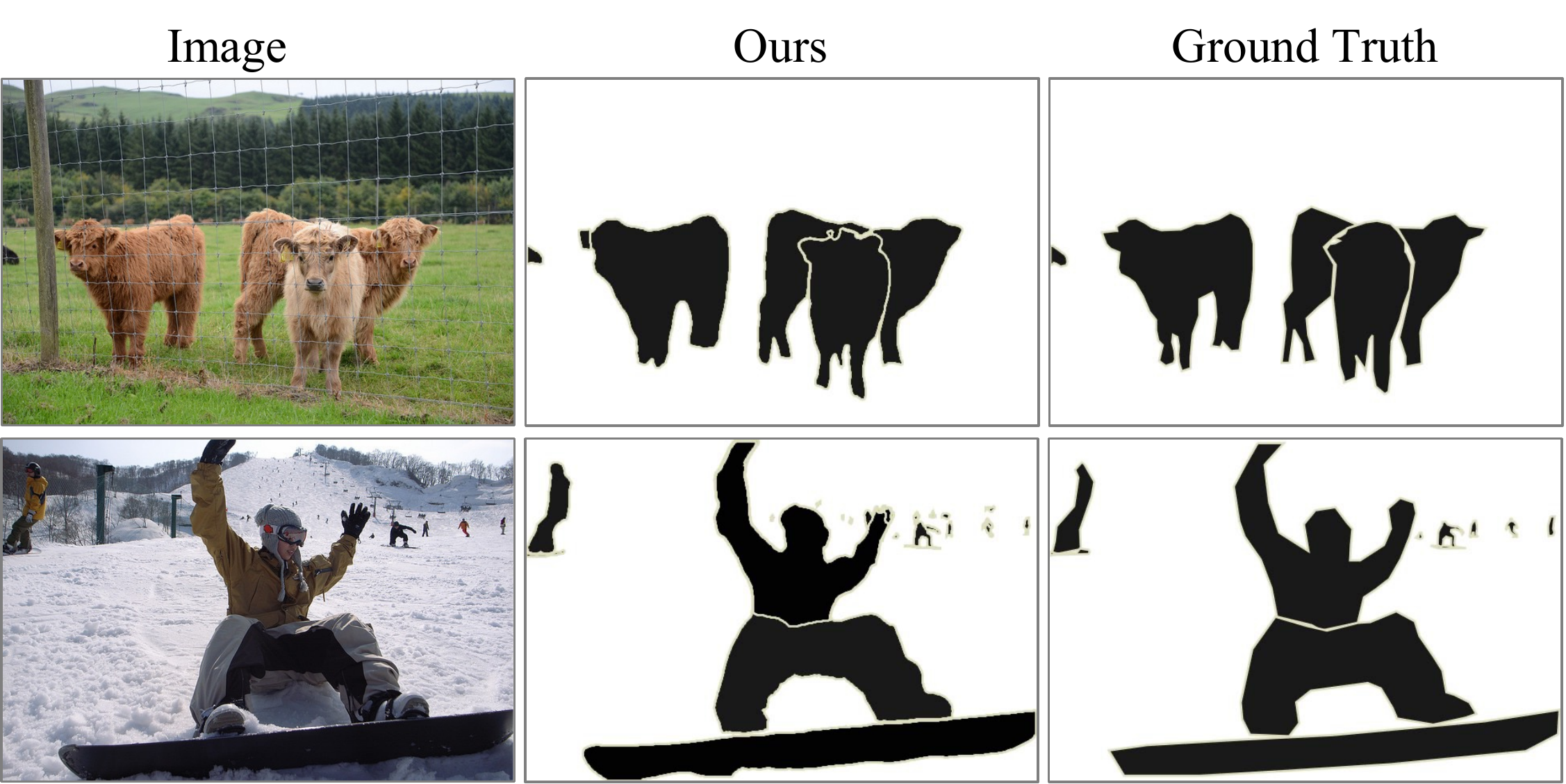}
	\vspace{-0.3cm}
	\caption{The predicted background masks by our approach can well exclude novel foreground objects of \texttt{cow} and \texttt{snowboard}.} 
	\vspace{-0.2cm}
	\label{fig:background_mask}
\end{figure}

In \figurename~\ref{fig:background_mask}, we visualize our generated background masks on unseen classes, \eg, \texttt{cow} and \texttt{snowboard}. As shown in Figure~\ref{fig:background_mask}, the foreground masks can be well segmented for both the seen and unseen classes. Because of the satisfactory generalization ability of mask proposal, we can generate a meaningful background mask for each image and produce a high-quality background representation.

\vspace{1mm}
\noindent\textbf{Unseen Cross-Entropy Loss.}
When introducing pseudo unseen labels generated from the seen-unseen similarity of semantic embeddings, the performance is significantly improved by 6.05\% HM-mAP and 5.12\% HM-Recall over baseline (see \tablename~\ref{tab:ablation_study} $\mathcal{L}^u$). This demonstrates that with the help of pseudo unseen labels, the feature extractor trained under the constraints of unseen categories can significantly alleviate bias towards seen classes issue and be well generalized to noval objects that have never show up in training.

\vspace{1mm}
\noindent\textbf{Generalization Ability of Instance Proposal.}
In Table~\ref{tab:generalization}, we test the category-agnostic mask proposal of unseen classes at different IoU thresholds, using the model trained on ``seen'' and the model trained on ``seen + unseen''. Training on ``seen'' achieves competitive results, demonstrating that Mask2Former can output masks for unseen categories when only trained with seen categories. 

\begin{table}[t]
\centering
\begin{minipage}[t]{0.49\textwidth}
\scriptsize
 \begin{minipage}[t]{0.4\textwidth}
  \centering
     \scriptsize
     \setlength\tabcolsep{3pt}
       \begin{tabular}{c|ccc} 
        \toprule
        \multirow{2}{*}{\makecell[c]{Training\\Categories}}&\multicolumn{3}{c}{AR@100}\\
          & 0.4 & 0.5 & 0.6 \\ \hline
        Seen &  78.5& 73.4 & 67.8 \\ 
        {Seen + Unseen} &82.1 & 78.4&73.1\\
        \bottomrule
         \end{tabular}
    \vspace{-3mm}
  \centering\makeatletter\def\@captype{table}\makeatother\caption{Instance proposal generalization ability.}\label{tab:generalization}
  \end{minipage}
  \hspace{6mm}
  \begin{minipage}[t]{0.45\textwidth}
   \centering
        \scriptsize
        \setlength\tabcolsep{3pt}
         \begin{tabular}{l|ccc}        
          \toprule
          Method  & mAP & AP50 & AP75  \\ \hline
          ZSI~(w2v)~\cite{zsi} & 0.008  & 0.009  & 0.008 \\ 
          \rowcolor{lightgray!24}\ours~(w2v) &4.670 &7.025  &4.816  \\ 
          \rowcolor{lightgray!24}\ours~(clip) &6.093 & 8.993  &6.279 \\ 
          \bottomrule
      \end{tabular}
      \vspace{-3mm}
     \makeatletter\def\@captype{table}\makeatother\caption{Cross-dataset results on ADE20k validation dataset.}\label{tab:transfer}
   \end{minipage}
\end{minipage}
\vspace{-5mm}
\end{table}

\begin{figure*}[t]
    \centering
	\includegraphics[width=0.96\textwidth]{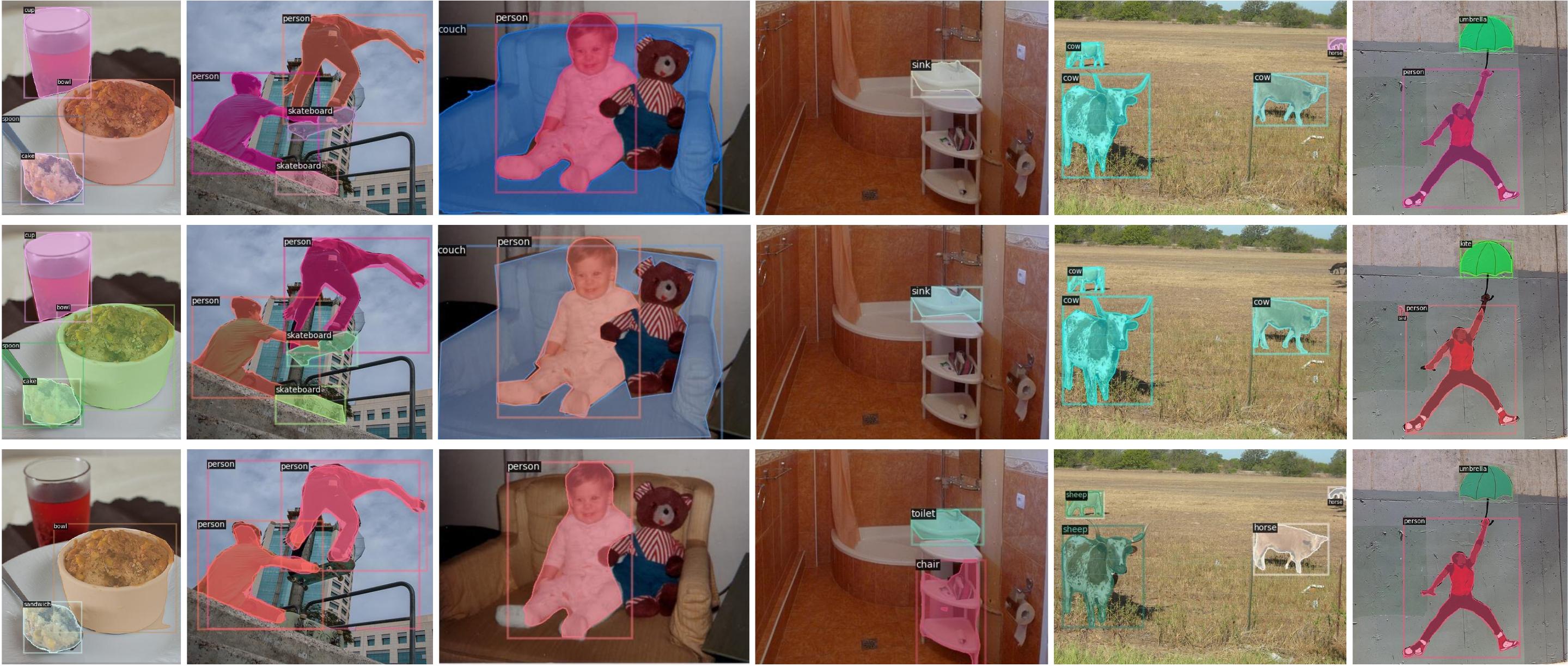}
	\vspace{-0.36cm}
	\caption{(Best viewed in color) From the 1st row to 3rd row are: ground truth, our results, and ZSI~\cite{zsi}, respectively. ZSI fails to classify most of the unseen objects, \eg, \texttt{cake} in the first image and \texttt{cow} in the fifth image. And some novel objects are missed by ZSI due to background confusion, \eg, \texttt{skateboard} in the second image and \texttt{couch} in the third image. The proposed approach \ours~shows much better results by classification debiasing and background disambiguation.} 
	\vspace{-0.36cm}
	\label{fig:demo}
\end{figure*}

\begin{table}[t]
\setlength\tabcolsep{4.2pt}
\scriptsize
  \begin{center}
  \begin{tabular}{cl|cccccc}
  \toprule
  & Method  &\multicolumn{2}{c}{Seen} & \multicolumn{2}{c}{Unseen} & \multicolumn{2}{c}{HM}\\
Split &   (text encoder)  &  mAP  & Recall & mAP  & Recall & mAP & Recall \\
  \hline
   \multirow{5}{*}{48/17}& ZSI (w2v)~\cite{zsi} &  43.04&  64.48  &\ \ 3.65 & 44.90 &\ \ 6.73 & 52.94 \\
   \cline{2-8}
   &\cellcolor{lightgray!20}\ours~(w2v) &\cellcolor{lightgray!20}{52.53}  &\cellcolor{lightgray!20}{75.66}  &\cellcolor{lightgray!20}{9.48}  &\cellcolor{lightgray!20}{37.93}  &\cellcolor{lightgray!20}{16.06} &\cellcolor{lightgray!20}{50.52}\\
      &\cellcolor{lightgray!20}\ours~(w2v-cp) &\cellcolor{lightgray!20}{51.75}  &\cellcolor{lightgray!20}{73.23}  &\cellcolor{lightgray!20}{10.58}  &\cellcolor{lightgray!20}{50.78}  &\cellcolor{lightgray!20}{17.56} &\cellcolor{lightgray!20}{59.97}\\
   &\cellcolor{lightgray!20}\ours~(clip) &\cellcolor{lightgray!20}\textbf{54.42} &   \cellcolor{lightgray!20}\textbf{76.22} &  \cellcolor{lightgray!20}{15.06}   &  \cellcolor{lightgray!20}{38.38}   & \cellcolor{lightgray!20}{23.59}    &  \cellcolor{lightgray!20}{51.06}\\
      &\cellcolor{lightgray!20}\ours~(clip-cp) &\cellcolor{lightgray!20}{54.12} &   \cellcolor{lightgray!20}{73.22} &  \cellcolor{lightgray!20}\textbf{15.82}   &  \cellcolor{lightgray!20}\textbf{53.53}   & \cellcolor{lightgray!20}\textbf{24.49}    &  \cellcolor{lightgray!20}\textbf{61.85}\\
  \arrayrulecolor{gray}\hline
   \multirow{5}{*}{65/15}& ZSI (w2v)~\cite{zsi} & 35.75 & 62.58 & 10.47 & 49.95 & 16.20 & 55.56\\
   \cline{2-8}
    &\cellcolor{lightgray!20}\ours~(w2v) &\cellcolor{lightgray!20}{38.49}  &\cellcolor{lightgray!20}{74.25}  &\cellcolor{lightgray!20}{13.12}  &\cellcolor{lightgray!20}{41.67}  & \cellcolor{lightgray!20}{19.57} &\cellcolor{lightgray!20}{53.38}\\
    &\cellcolor{lightgray!20}\ours~(w2v-cp) &\cellcolor{lightgray!20}{37.32} &   \cellcolor{lightgray!20}{70.43} &  \cellcolor{lightgray!20}{15.39}   &  \cellcolor{lightgray!20}{58.64}   & \cellcolor{lightgray!20}{21.79}    &  \cellcolor{lightgray!20}{63.99}\\
 &\cellcolor{lightgray!20}\ours~(clip) &\cellcolor{lightgray!20}\textbf{41.18}  &\cellcolor{lightgray!20}\textbf{74.94}  &\cellcolor{lightgray!20}{20.22}  &\cellcolor{lightgray!20}{46.01}  & \cellcolor{lightgray!20}{27.13} &\cellcolor{lightgray!20}{57.01}\\
   &\cellcolor{lightgray!20}\ours~(clip-cp) &\cellcolor{lightgray!20}40.90 &\cellcolor{lightgray!20}{71.41}  &\cellcolor{lightgray!20}\textbf{ 21.91}&\cellcolor{lightgray!20}\textbf{65.72}  &\cellcolor{lightgray!20}\textbf{28.54} &\cellcolor{lightgray!20}\textbf{68.45}\\
 \arrayrulecolor{black}\bottomrule
  \end{tabular}
\vspace{-3mm}
\caption{Results on GZSIS. ``cp'' denotes copy-paste strategy of ZSI~\cite{zsi}, \ie, sharing instances between seen and unseen groups.}\label{tab:SOTA_GZSI}
\vspace{-5mm}
\end{center}
\end{table}

\begin{table}[t]
\setlength\tabcolsep{10pt}
\scriptsize
  \begin{center}
  \begin{tabular}{cl|cccc}
\toprule
  &  Method &\multicolumn{3}{c}{Recall@100} & mAP\\
Split &   (text encoder)  &  0.4  & 0.5 & 0.6 & 0.5 \\
  \hline
   \multirow{3}{*}{48/17}& ZSI (w2v)~\cite{zsi} &  50.3 & 44.9 & 38.7 &\ \  9.0\\
   \cline{2-6}
   &\cellcolor{lightgray!20}\ours~(w2v) &\cellcolor{lightgray!20}{60.0} &\cellcolor{lightgray!20}{55.9} &\cellcolor{lightgray!20}{50.8} &\cellcolor{lightgray!20}{16.1}\\
   &\cellcolor{lightgray!20}\ours~(clip) &\cellcolor{lightgray!20}\textbf{65.5} &\cellcolor{lightgray!20}\textbf{61.4} &\cellcolor{lightgray!20}\textbf{55.9} &\cellcolor{lightgray!20}\textbf{21.7}\\
  \arrayrulecolor{gray}\hline
   \multirow{3}{*}{65/15}& ZSI (w2v)~\cite{zsi} & 55.8 & 50.0 & 42.9 & 10.5\\
   \cline{2-6}
   &\cellcolor{lightgray!20}\ours~(w2v) &\cellcolor{lightgray!20}{68.5} &\cellcolor{lightgray!20}{65.1} &\cellcolor{lightgray!20}{60.6} &\cellcolor{lightgray!20}{16.9}\\
   &\cellcolor{lightgray!20}\ours~(clip) &\cellcolor{lightgray!20}\textbf{73.3} &\cellcolor{lightgray!20}\textbf{69.7} &\cellcolor{lightgray!20}\textbf{64.9} &\cellcolor{lightgray!20}\textbf{23.7}\\
  \arrayrulecolor{black}\bottomrule
  \end{tabular}
\vspace{-3mm}
\caption{Results on ZSIS.}\label{tab:SOTA_ZSIS}
\vspace{-9mm}
\end{center}
\end{table}

\subsection{Transfer to Other Dataset}
\ours~model trained on COCO can be transferred to other instance segmentation datasets like ADE20k~\cite{zhou2017scene} via replacing semantic embeddings of our input-conditional classifier. We input the semantic embeddings of ADE20K classes as $Q$ to our input-conditional classifier and testing our results on ADE20K, as shown in \tablename~\ref{tab:transfer}. 
Our method demonstrates good generalization ability on cross-dataset testing and significantly outperforms ZSI~\cite{zsi}. In ZSI~\cite{zsi}, there are some classifier parameters related to the dataset category, making it impossible to transfer to other datasets.

\subsection{Comparison with State-of-the-Art Methods}
In \tablename~\ref{tab:SOTA_ZSIS} and \tablename~\ref{tab:SOTA_GZSI}, we follow the experimental settings in ZSI~\cite{zsi} to report our results on both the Zero-Shot Instance Segmentation (ZSIS) and Generalized Zero-Shot Instance Segmentation (GZSIS) tasks. The proposed \ours~exceeds ZSI by a large margin, \eg, our model with CLIP~\cite{CLIP} as text encoder outperforms ZSI by 16.86\% HM-mAP under the 48/17 split and 10.93\% H-mAP under the 65/15 split. We also report our results using copy-paste strategy of ZSI, marked with ``cp''. The ``cp'' strategy significantly improves recall performance for unseen classes but decreases precision since it brings many false positives.
To further evaluate the superiority of our method, we conduct a model complexity comparison with ZSI. 
 The \#parameters/FLOPs of our \ours~and ZSI~\cite{zsi} are 45.737M/227.7G and 69.6M/569.3G, 
\ours~is dramatically more efficient than ZSI, thanks to our efficient mask proposal network and lightweight component design.

In \figurename~\ref{fig:demo}, we present a qualitative comparison with ZSI~\cite{zsi} for both seen and unseen classes on COCO under the 48/17 split. 
ZSI fails to classify most of the unseen objects. For example, ZSI identifies \texttt{cow} as \texttt{horse}. By contrast, our approach outputs more accurate instance labels for both seen and unseen categories and more precise mask predictions. Besides, our method successfully segments the objects missed by ZSI due to background ambiguation, like \texttt{couch} in the 3rd column, which demonstrates the effectiveness of our background disambiguation.

\begin{table}[t]
\setlength\tabcolsep{3.96pt}
\scriptsize
  \begin{center}
  \begin{tabular}{cl|cccccc}
  \toprule
  &  &\multicolumn{2}{c}{Seen} & \multicolumn{2}{c}{Unseen} & \multicolumn{2}{c}{HM}\\
Split &   Method  &  mAP  & Recall & mAP  & Recall & mAP & Recall \\
  \hline
  
   \multirow{6}{*}{48/17}& DSES~\cite{bansal2018zero} & - & 15.02 & - & 15.32 & - & 15.17 \\
   & PL~\cite{rahman2020improved}& 35.92 & 38.24 &\ \ 4.12 & 26.32 &\ \ 7.39 & 31.18 \\
   &BLC~\cite{zheng2020BLC} & 42.10 & 57.56 &\ \ 4.50 & 46.39 &\ \ 8.20 & 51.37 \\
   & ZSI~\cite{zsi} & {46.51} &{70.76} &\ \  {4.83} & {53.85} &\ \ {8.75} & {61.16}\\
   \cline{2-8}
   &\cellcolor{lightgray!20}\ours~(w2v) &\cellcolor{lightgray!20}{52.30}  &\cellcolor{lightgray!20}{76.89}  &\cellcolor{lightgray!20}{ 9.46}  &\cellcolor{lightgray!20}{37.28}  &\cellcolor{lightgray!20}{16.02} &\cellcolor{lightgray!20}{50.21}\\
    & \cellcolor{lightgray!20}\ours~(w2v-cp) &\cellcolor{lightgray!20}{51.31}  &\cellcolor{lightgray!20}{72.04}  &\cellcolor{lightgray!20}{10.55}  &\cellcolor{lightgray!20}{54.14}  &\cellcolor{lightgray!20} {17.50} &\cellcolor{lightgray!20}{62.01}\\
   &\cellcolor{lightgray!20}\ours~(clip) &\cellcolor{lightgray!20}\textbf{54.47}  &\cellcolor{lightgray!20}\textbf{77.52}  &\cellcolor{lightgray!20}{14.67}  &\cellcolor{lightgray!20}{38.09}  & \cellcolor{lightgray!20}{23.12} &\cellcolor{lightgray!20}{51.08}\\

   &\cellcolor{lightgray!20}\ours~(clip-cp) &\cellcolor{lightgray!20}{54.14}  &\cellcolor{lightgray!20}{74.09}  &\cellcolor{lightgray!20}\textbf{15.45}  &\cellcolor{lightgray!20}\textbf{54.19}  & \cellcolor{lightgray!20}\textbf{24.05} &\cellcolor{lightgray!20}\textbf{62.60}\\
  \arrayrulecolor{gray}\hline
   \multirow{5}{*}{65/15}& PL~\cite{rahman2020improved}& 34.07 & 36.38 & 12.40 & 37.16 & 18.18 & 36.76 \\
   &BLC~\cite{zheng2020BLC} & 36.00 & 56.39 & 13.10 & 51.65 & 19.20 & 53.92\\
   & ZSI~\cite{zsi} & 38.68 & 67.11 & 13.60 & 58.93 & 20.13 & 62.76 \\
   \cline{2-8}
    &\cellcolor{lightgray!20}\ours~(w2v) &\cellcolor{lightgray!20}{38.71}  &\cellcolor{lightgray!20}{74.25}  &\cellcolor{lightgray!20}{13.00}  &\cellcolor{lightgray!20}{41.41}  &\cellcolor{lightgray!20} {19.46} &\cellcolor{lightgray!20}{53.16}\\
        &\cellcolor{lightgray!20}\ours~(w2v-cp) &\cellcolor{lightgray!20}{39.32}  &\cellcolor{lightgray!20}{70.43}  &\cellcolor{lightgray!20}{15.39}  &\cellcolor{lightgray!20}{63.64}  &\cellcolor{lightgray!20} {22.12} &\cellcolor{lightgray!20}{66.86}\\
 &\cellcolor{lightgray!20}\ours~(clip) &\cellcolor{lightgray!20}\textbf{40.51}  &\cellcolor{lightgray!20}\textbf{74.64}  &\cellcolor{lightgray!20}{20.23}  &\cellcolor{lightgray!20}{46.33}  &\cellcolor{lightgray!20}{26.99} &\cellcolor{lightgray!20}{57.17}\\

  &\cellcolor{lightgray!20}\ours~(clip-cp) &\cellcolor{lightgray!20}{40.23}  &\cellcolor{lightgray!20}{70.98}  &\cellcolor{lightgray!20}\textbf{21.84}  &\cellcolor{lightgray!20}\textbf{66.65}  & \cellcolor{lightgray!20}\textbf{28.31} &\cellcolor{lightgray!20}\textbf{68.75}\\
   \arrayrulecolor{black}
 \bottomrule
  \end{tabular}
\vspace{-3mm}
\caption{Results on GZSD. Previous methods all use word2vector.}\label{tab:SOTA_GZSD}
\vspace{-7mm}
\end{center}
\end{table}

\begin{table}[t]
\setlength\tabcolsep{9pt}
\scriptsize
  \begin{center}
  \begin{tabular}{cl|cccc}
\toprule
  &  &\multicolumn{3}{c}{Recall@100} & mAP\\
Split &   Method  &  0.4  & 0.5 & 0.6 & 0.5 \\
  \hline

\multirow{10}{*}{48/17}& SB~\cite{bansal2018zero} & 34.46 & 22.14 & 11.31 &\ \ 0.32 \\
   &DSES~\cite{bansal2018zero}  &40.23 & 27.19 & 13.63 &\ \ 0.54\\
      &TD~\cite{li2019zero} &  45.50 & 34.30 & 18.10 & - \\
   & PL~\cite{rahman2020improved}& - & 43.59 & - & 10.10 \\
      & Gtnet~\cite{zhao2020gtnet} &  47.30 & 44.60 & 35.50 & - \\ 
   & DELO~\cite{zhu2020don} &  - & 33.50 & - &\ \ 7.60 \\ 
   &BLC~\cite{zheng2020BLC}  & 49.63 & 46.39 & 41.86 &\ \ 9.90\\
   & ZSI~\cite{zsi} & {57.40} &  {53.90} & {48.30} & {11.40}\\
   \cline{2-6}
   &\cellcolor{lightgray!20}\ours~(w2v) &\cellcolor{lightgray!20}{60.00} &\cellcolor{lightgray!20}{56.10} &\cellcolor{lightgray!20}{52.00} &\cellcolor{lightgray!20}{16.30}\\
   &\cellcolor{lightgray!20}\ours~(clip) &\cellcolor{lightgray!20}\textbf{65.70} &\cellcolor{lightgray!20}\textbf{61.70} &\cellcolor{lightgray!20}\textbf{57.70} &\cellcolor{lightgray!20}\textbf{21.40}\\
  \arrayrulecolor{gray}\hline
   \multirow{5}{*}{65/15}& PL~\cite{rahman2020improved}&  - & 37.72 & - & 12.40 \\
   &BLC~\cite{zheng2020BLC} &  54.18 & 51.65 & 47.86 & 13.10\\
   & ZSI~\cite{zsi} & 61.90& 58.90 & 54.40 & 13.60\\
   \cline{2-6}
   &\cellcolor{lightgray!20}\ours(w2v) &\cellcolor{lightgray!20}{69.10} &\cellcolor{lightgray!20}{66.20} &\cellcolor{lightgray!20}{62.30} &\cellcolor{lightgray!20}{16.80}\\
   &\cellcolor{lightgray!20}\ours(clip) &\cellcolor{lightgray!20}\textbf{73.90} &\cellcolor{lightgray!20}\textbf{70.70} &\cellcolor{lightgray!20}\textbf{66.60} &\cellcolor{lightgray!20}\textbf{23.50}\\
  \arrayrulecolor{black}\bottomrule
  \end{tabular}
\vspace{-3mm}
\caption{Results on ZSD. Previous methods all use word2vector.}\label{tab:SOTA_ZSD}
\vspace{-8mm}
\end{center}
\end{table}

We also report our results on Zero-Shot Detection (ZSD) in \tablename~\ref{tab:SOTA_ZSD} and Generalized Zero-Shot Detection (GZSD) in \tablename~\ref{tab:SOTA_GZSD}. We do not use bounding box regression but simply produce bounding box from our masks, and achieves new state-of-the-art performance on ZSD and GZSD. The above experiments and analysis all demonstrate the effectiveness and efficiency of our \ours~on Zero-shot instance segmentation and detection tasks.
\vspace{-2mm}
\section{Conclusion}
We propose \ours~with semantic-promoted debiasing and background disambiguation to address the two key challenges in zero-shot instance segmentation, \ie, bias issue and background ambiguation. To alleviate the bias issue, we introduce a semantic-constrained feature training strategy to utilize semantic knowledge of unseen classes and propose an input-conditional classifier to dynamically produce image-specific prototypes for classification. We discuss the background confusion and build an image-adaptive background prototype to better capture discriminative background clues. We achieve new state-of-the-art results on zero-shot instance segmentation and detection.

\footnotesize{\noindent\textbf{Acknowledgement} This work was partially supported by National Natural Science Foundation of China (No.62173302).}

\begin{spacing}{0.96}
{\small
\bibliographystyle{ieee_fullname}
\bibliography{egbib}

\begin{thebibliography}{10}\itemsep=-1pt

\bibitem{akata2015evaluation}
Zeynep Akata, Scott Reed, Daniel Walter, Honglak Lee, and Bernt Schiele.
\newblock Evaluation of output embeddings for fine-grained image
  classification.
\newblock In {\em CVPR}, 2015.

\bibitem{bansal2018zero}
Ankan Bansal, Karan Sikka, Gaurav Sharma, Rama Chellappa, and Ajay Divakaran.
\newblock Zero-shot object detection.
\newblock In {\em ECCV}, 2018.

\bibitem{bhattacharjee2019autoencoder}
Supritam Bhattacharjee, Devraj Mandal, and Soma Biswas.
\newblock Autoencoder based novelty detection for generalized zero shot
  learning.
\newblock In {\em ICIP}, 2019.

\bibitem{bolya2019yolact}
Daniel Bolya, Chong Zhou, Fanyi Xiao, and Yong~Jae Lee.
\newblock Yolact: Real-time instance segmentation.
\newblock In {\em ICCV}, 2019.

\bibitem{ZS3}
Maxime Bucher, Tuan-Hung Vu, Matthieu Cord, and Patrick P{\'e}rez.
\newblock Zero-shot semantic segmentation.
\newblock {\em NeurIPS}, 32, 2019.

\bibitem{detr}
Nicolas Carion, Francisco Massa, Gabriel Synnaeve, Nicolas Usunier, Alexander
  Kirillov, and Sergey Zagoruyko.
\newblock End-to-end object detection with transformers.
\newblock In {\em ECCV}, 2020.

\bibitem{changpinyo2020classifier}
Soravit Changpinyo, Wei-Lun Chao, Boqing Gong, and Fei Sha.
\newblock Classifier and exemplar synthesis for zero-shot learning.
\newblock {\em IJCV}, 128(1), 2020.

\bibitem{chao2016empirical}
Wei-Lun Chao, Soravit Changpinyo, Boqing Gong, and Fei Sha.
\newblock An empirical study and analysis of generalized zero-shot learning for
  object recognition in the wild.
\newblock In {\em ECCV}, 2016.

\bibitem{mask2former}
Bowen Cheng, Ishan Misra, Alexander~G Schwing, Alexander Kirillov, and Rohit
  Girdhar.
\newblock Masked-attention mask transformer for universal image segmentation.
\newblock In {\em CVPR}, 2022.

\bibitem{SIGN}
Jiaxin Cheng, Soumyaroop Nandi, Prem Natarajan, and Wael Abd-Almageed.
\newblock Sign: Spatial-information incorporated generative network for
  generalized zero-shot semantic segmentation.
\newblock In {\em ICCV}, 2021.

\bibitem{das2019zero}
Debasmit Das and CS~George Lee.
\newblock Zero-shot image recognition using relational matching, adaptation and
  calibration.
\newblock In {\em IJCNN}, 2019.

\bibitem{demirel2017attributes2classname}
Berkan Demirel, Ramazan Gokberk~Cinbis, and Nazli Ikizler-Cinbis.
\newblock Attributes2classname: A discriminative model for attribute-based
  unsupervised zero-shot learning.
\newblock In {\em ICCV}, 2017.

\bibitem{ding2018context}
Henghui Ding, Xudong Jiang, Bing Shuai, Ai~Qun Liu, and Gang Wang.
\newblock Context contrasted feature and gated multi-scale aggregation for
  scene segmentation.
\newblock In {\em CVPR}, 2018.

\bibitem{VLT}
Henghui Ding, Chang Liu, Suchen Wang, and Xudong Jiang.
\newblock Vision-language transformer and query generation for referring
  segmentation.
\newblock In {\em ICCV}, 2021.

\bibitem{ding2022vlt}
Henghui Ding, Chang Liu, Suchen Wang, and Xudong Jiang.
\newblock {VLT}: Vision-language transformer and query generation for referring
  segmentation.
\newblock {\em IEEE TPAMI}, 2023.

\bibitem{ZegFormer}
Jian Ding, Nan Xue, Gui-Song Xia, and Dengxin Dai.
\newblock Decoupling zero-shot semantic segmentation.
\newblock In {\em CVPR}, 2022.

\bibitem{dinu2014improving}
Georgiana Dinu, Angeliki Lazaridou, and Marco Baroni.
\newblock Improving zero-shot learning by mitigating the hubness problem.
\newblock {\em arXiv preprint arXiv:1412.6568}, 2014.

\bibitem{felix2019generalised}
Rafael Felix, Ben Harwood, Michele Sasdelli, and Gustavo Carneiro.
\newblock Generalised zero-shot learning with domain classification in a joint
  semantic and visual space.
\newblock In {\em DICTA}, 2019.

\bibitem{frome2013devise}
Andrea Frome, Greg~S Corrado, Jon Shlens, Samy Bengio, Jeff Dean, Marc'Aurelio
  Ranzato, and Tomas Mikolov.
\newblock Devise: A deep visual-semantic embedding model.
\newblock {\em NeurIPS}, 26, 2013.

\bibitem{ghiasi2022scaling}
Golnaz Ghiasi, Xiuye Gu, Yin Cui, and Tsung-Yi Lin.
\newblock Scaling open-vocabulary image segmentation with image-level labels.
\newblock In {\em ECCV}, 2022.

\bibitem{VILD}
Xiuye Gu, Tsung-Yi Lin, Weicheng Kuo, and Yin Cui.
\newblock Open-vocabulary object detection via vision and language knowledge
  distillation.
\newblock {\em arXiv preprint arXiv:2104.13921}, 2021.

\bibitem{CaGNet}
Zhangxuan Gu, Siyuan Zhou, Li Niu, Zihan Zhao, and Liqing Zhang.
\newblock Context-aware feature generation for zero-shot semantic segmentation.
\newblock In {\em ACM MM}, 2020.

\bibitem{guo2019dual}
Yuchen Guo, Guiguang Ding, Jungong Han, Xiaohan Ding, Sicheng Zhao, Zheng Wang,
  Chenggang Yan, and Qionghai Dai.
\newblock Dual-view ranking with hardness assessment for zero-shot learning.
\newblock In {\em AAAI}, volume~33, 2019.

\bibitem{maskrcnn}
Kaiming He, Georgia Gkioxari, Piotr Doll{\'a}r, and Ross Girshick.
\newblock Mask r-cnn.
\newblock In {\em ICCV}, 2017.

\bibitem{he2016deep}
Kaiming He, Xiangyu Zhang, Shaoqing Ren, and Jian Sun.
\newblock Deep residual learning for image recognition.
\newblock In {\em CVPR}, 2016.

\bibitem{PADing}
Shuting He, Henghui Ding, and Wei Jiang.
\newblock Primitive generation and semantic-related alignment for universal
  zero-shot segmentation.
\newblock In {\em CVPR}, 2023.

\bibitem{FZShot3D}
Shuting He, Xudong Jiang, Wei Jiang, and Henghui Ding.
\newblock Prototype adaption and projection for few- and zero-shot 3d point
  cloud semantic segmentation.
\newblock {\em IEEE TIP}, 2023.

\bibitem{hu2020uncertainty}
Ping Hu, Stan Sclaroff, and Kate Saenko.
\newblock Uncertainty-aware learning for zero-shot semantic segmentation.
\newblock {\em NeurIPS}, 33, 2020.

\bibitem{hu2023suppressing}
Zhengdong Hu, Yifan Sun, and Yi Yang.
\newblock Suppressing the heterogeneity: A strong feature extractor for
  few-shot segmentation.
\newblock In {\em ICLR}, 2023.

\bibitem{huynh2022open}
Dat Huynh, Jason Kuen, Zhe Lin, Jiuxiang Gu, and Ehsan Elhamifar.
\newblock Open-vocabulary instance segmentation via robust cross-modal
  pseudo-labeling.
\newblock In {\em CVPR}, 2022.

\bibitem{GumbelSoftmax}
Eric Jang, Shixiang Gu, and Ben Poole.
\newblock Categorical reparametrization with gumble-softmax.
\newblock In {\em ICLR}, 2017.

\bibitem{kankuekul2012online}
Pichai Kankuekul, Aram Kawewong, Sirinart Tangruamsub, and Osamu Hasegawa.
\newblock Online incremental attribute-based zero-shot learning.
\newblock In {\em CVPR}, 2012.

\bibitem{OLN}
Dahun Kim, Tsung-Yi Lin, Anelia Angelova, In~So Kweon, and Weicheng Kuo.
\newblock Learning open-world object proposals without learning to classify.
\newblock {\em IEEE RA-L}, 2022.

\bibitem{lampert2009learning}
Christoph~H Lampert, Hannes Nickisch, and Stefan Harmeling.
\newblock Learning to detect unseen object classes by between-class attribute
  transfer.
\newblock In {\em CVPR}, 2009.

\bibitem{lampert2013attribute}
Christoph~H Lampert, Hannes Nickisch, and Stefan Harmeling.
\newblock Attribute-based classification for zero-shot visual object
  categorization.
\newblock {\em TPAMI}, 36(3), 2013.

\bibitem{li2022languagedriven}
Boyi Li, Kilian~Q Weinberger, Serge Belongie, Vladlen Koltun, and Rene Ranftl.
\newblock Language-driven semantic segmentation.
\newblock In {\em ICLR}, 2022.

\bibitem{CSRL}
Peike Li, Yunchao Wei, and Yi Yang.
\newblock Consistent structural relation learning for zero-shot segmentation.
\newblock {\em NeurIPS}, 33, 2020.

\bibitem{TransformerSurvey}
Xiangtai Li, Henghui Ding, Wenwei Zhang, Haobo Yuan, Jiangmiao Pang, Guangliang
  Cheng, Kai Chen, Ziwei Liu, and Chen~Change Loy.
\newblock Transformer-based visual segmentation: A survey.
\newblock {\em arXiv:2304.09854}, 2023.

\bibitem{li2018deep}
Yan Li, Zhen Jia, Junge Zhang, Kaiqi Huang, and Tieniu Tan.
\newblock Deep semantic structural constraints for zero-shot learning.
\newblock In {\em AAAI}, 2018.

\bibitem{li2019zero}
Zhihui Li, Lina Yao, Xiaoqin Zhang, Xianzhi Wang, Salil Kanhere, and Huaxiang
  Zhang.
\newblock Zero-shot object detection with textual descriptions.
\newblock In {\em AAAI}, 2019.

\bibitem{ms_coco}
Tsung-Yi Lin, Michael Maire, Serge Belongie, James Hays, Pietro Perona, Deva
  Ramanan, Piotr Doll{\'a}r, and C~Lawrence Zitnick.
\newblock Microsoft coco: Common objects in context.
\newblock In {\em ECCV}, 2014.

\bibitem{GRES}
Chang Liu, Henghui Ding, and Xudong Jiang.
\newblock {GRES}: Generalized referring expression segmentation.
\newblock In {\em CVPR}, 2023.

\bibitem{M3Att}
Chang Liu, Henghui Ding, Yulun Zhang, and Xudong Jiang.
\newblock Multi-modal mutual attention and iterative interaction for referring
  image segmentation.
\newblock {\em IEEE TIP}, 2023.

\bibitem{liu2022instance}
Chang Liu, Xudong Jiang, and Henghui Ding.
\newblock Instance-specific feature propagation for referring segmentation.
\newblock {\em IEEE TMM}, 2022.

\bibitem{lu2021simpler}
Zhihe Lu, Sen He, Xiatian Zhu, Li Zhang, Yi-Zhe Song, and Tao Xiang.
\newblock Simpler is better: Few-shot semantic segmentation with classifier
  weight transformer.
\newblock In {\em ICCV}, 2021.

\bibitem{lv2020learning}
Fengmao Lv, Haiyang Liu, Yichen Wang, Jiayi Zhao, and Guowu Yang.
\newblock Learning unbiased zero-shot semantic segmentation networks via
  transductive transfer.
\newblock {\em IEEE SPL}, 27, 2020.

\bibitem{word2vector}
Tomas Mikolov, Kai Chen, Greg Corrado, and Jeffrey Dean.
\newblock Efficient estimation of word representations in vector space.
\newblock {\em arXiv preprint arXiv:1301.3781}, 2013.

\bibitem{mikolov2013distributed}
Tomas Mikolov, Ilya Sutskever, Kai Chen, Greg~S Corrado, and Jeff Dean.
\newblock Distributed representations of words and phrases and their
  compositionality.
\newblock In {\em NeurIPS}, 2013.

\bibitem{palatucci2009zero}
Mark Palatucci, Dean Pomerleau, Geoffrey~E Hinton, and Tom~M Mitchell.
\newblock Zero-shot learning with semantic output codes.
\newblock {\em NeurIPS}, 22, 2009.

\bibitem{pastore2021closer}
Giuseppe Pastore, Fabio Cermelli, Yongqin Xian, Massimiliano Mancini, Zeynep
  Akata, and Barbara Caputo.
\newblock A closer look at self-training for zero-label semantic segmentation.
\newblock In {\em CVPR}, 2021.

\bibitem{CLIP}
Alec Radford, Jong~Wook Kim, Chris Hallacy, Aditya Ramesh, Gabriel Goh,
  Sandhini Agarwal, Girish Sastry, Amanda Askell, Pamela Mishkin, Jack Clark,
  et~al.
\newblock Learning transferable visual models from natural language
  supervision.
\newblock In {\em ICML}, 2021.

\bibitem{rahman2020improved}
Shafin Rahman, Salman Khan, and Nick Barnes.
\newblock Improved visual-semantic alignment for zero-shot object detection.
\newblock {\em AAAI}, 2020.

\bibitem{scheirer2012toward}
Walter~J Scheirer, Anderson de Rezende~Rocha, Archana Sapkota, and Terrance~E
  Boult.
\newblock Toward open set recognition.
\newblock {\em TPAMI}, 35(7), 2012.

\bibitem{socher2013zero}
Richard Socher, Milind Ganjoo, Hamsa Sridhar, Osbert Bastani, Christopher~D
  Manning, and Andrew~Y Ng.
\newblock Zero-shot learning through cross-modal transfer.
\newblock {\em arXiv}, 2013.

\bibitem{tsne}
Laurens Van~der Maaten and Geoffrey Hinton.
\newblock Visualizing data using t-sne.
\newblock {\em JMLR}, 9(11), 2008.

\bibitem{vaswani2017attention}
Ashish Vaswani, Noam Shazeer, Niki Parmar, Jakob Uszkoreit, Llion Jones,
  Aidan~N Gomez, {\L}ukasz Kaiser, and Illia Polosukhin.
\newblock Attention is all you need.
\newblock In {\em NeurIPS}, 2017.

\bibitem{solov1}
Xinlong Wang, Tao Kong, Chunhua Shen, Yuning Jiang, and Lei Li.
\newblock Solo: Segmenting objects by locations.
\newblock In {\em ECCV}, 2020.

\bibitem{wu2023betrayed}
Jianzong Wu, Xiangtai Li, Henghui Ding, Xia Li, Guangliang Cheng, Yunhai Tong,
  and Chen~Change Loy.
\newblock Betrayed by captions: Joint caption grounding and generation for open
  vocabulary instance segmentation.
\newblock {\em arXiv:2301.00805}, 2023.

\bibitem{SPNet}
Yongqin Xian, Subhabrata Choudhury, Yang He, Bernt Schiele, and Zeynep Akata.
\newblock Semantic projection network for zero-and few-label semantic
  segmentation.
\newblock In {\em CVPR}, 2019.

\bibitem{xian2018zero}
Yongqin Xian, Christoph~H Lampert, Bernt Schiele, and Zeynep Akata.
\newblock Zero-shot learning-a comprehensive evaluation of the good, the bad
  and the ugly.
\newblock {\em TPAMI}, 2018.

\bibitem{zsseg_baseline}
Mengde Xu, Zheng Zhang, Fangyun Wei, Yutong Lin, Yue Cao, Han Hu, and Xiang
  Bai.
\newblock A simple baseline for zero-shot semantic segmentation with
  pre-trained vision-language model.
\newblock {\em arXiv preprint arXiv:2112.14757}, 2021.

\bibitem{yang2022lavt}
Zhao Yang, Jiaqi Wang, Yansong Tang, Kai Chen, Hengshuang Zhao, and Philip~HS
  Torr.
\newblock Lavt: Language-aware vision transformer for referring image
  segmentation.
\newblock In {\em CVPR}, 2022.

\bibitem{yu2013designing}
Felix~X Yu, Liangliang Cao, Rogerio~S Feris, John~R Smith, and Shih-Fu Chang.
\newblock Designing category-level attributes for discriminative visual
  recognition.
\newblock In {\em CVPR}, 2013.

\bibitem{ovdetr}
Yuhang Zang, Wei Li, Kaiyang Zhou, Chen Huang, and Chen~Change Loy.
\newblock Open-vocabulary detr with conditional matching.
\newblock In {\em ECCV}, 2022.

\bibitem{ovrcnn}
Alireza Zareian, Kevin~Dela Rosa, Derek~Hao Hu, and Shih-Fu Chang.
\newblock Open-vocabulary object detection using captions.
\newblock In {\em CVPR}, 2021.

\bibitem{ding_iccv21}
Hui Zhang and Henghui Ding.
\newblock Prototypical matching and open set rejection for zero-shot semantic
  segmentation.
\newblock In {\em ICCV}, 2021.

\bibitem{zhao2020gtnet}
Shizhen Zhao, Changxin Gao, Yuanjie Shao, Lerenhan Li, Changqian Yu, Zhong Ji,
  and Nong Sang.
\newblock Gtnet: Generative transfer network for zero-shot object detection.
\newblock {\em arXiv preprint arXiv:2001.06812}, 2020.

\bibitem{BLC}
Ye Zheng, Ruoran Huang, Chuanqi Han, Xi Huang, and Li Cui.
\newblock Background learnable cascade for zero-shot object detection.
\newblock In {\em ACCV}, 2020.

\bibitem{zheng2020BLC}
Ye Zheng, Ruoran Huang, Chuanqi Han, Xi Huang, and Li Cui.
\newblock Background learnable cascade for zero-shot object detection.
\newblock {\em arXiv preprint arXiv:2010.04502}, 2020.

\bibitem{zsi}
Ye Zheng, Jiahong Wu, Yongqiang Qin, Faen Zhang, and Li Cui.
\newblock Zero-shot instance segmentation.
\newblock In {\em CVPR}, 2021.

\bibitem{zhou2017scene}
Bolei Zhou, Hang Zhao, Xavier Puig, Sanja Fidler, Adela Barriuso, and Antonio
  Torralba.
\newblock Scene parsing through ade20k dataset.
\newblock In {\em CVPR}, 2017.

\bibitem{zhu2020don}
Pengkai Zhu, Hanxiao Wang, and Venkatesh Saligrama.
\newblock Don't even look once: Synthesizing features for zero-shot detection.
\newblock In {\em CVPR}, 2020.

\end{thebibliography}
}
\end{spacing}

\end{document}